\documentclass[10pt,twocolumn,letterpaper]{article}

\usepackage{wacv}
\usepackage{times}
\usepackage{epsfig}
\usepackage{graphicx}
\usepackage{amsmath}
\usepackage{amssymb}
\usepackage{booktabs}
\usepackage{hhline}
\usepackage[ruled]{algorithm2e}
\usepackage{amsmath}
\usepackage[bottom]{footmisc}
\usepackage{caption,stackengine}
\usepackage[accsupp]{axessibility}
\usepackage{authblk}

\def\wacvPaperID{1099} 

\wacvapplicationstrack 

\wacvfinalcopy 

\ifwacvfinal
\def\assignedStartPage{3} 
\fi

\usepackage{pifont}
\usepackage{makecell}

\newcommand{\xmark}{\text{\ding{55}}}
\ifwacvfinal
\usepackage[breaklinks=true,bookmarks=false]{hyperref}
\else
\usepackage[pagebackref=true,breaklinks=true,colorlinks,bookmarks=false]{hyperref}
\fi

\ifwacvfinal
\else
\fi

\SetKwInput{KwInput}{Input}                
\SetKwInput{KwOutput}{Output} 
\newsavebox\CBox 
\def\textBF#1{\sbox\CBox{#1}\resizebox{\wd\CBox}{\ht\CBox}{\textbf{#1}}}

\usepackage{footnote}
\makesavenoteenv{tabular}

\makeatletter
\newcommand{\printfnsymbol}[1]{%
  \textsuperscript{\@fnsymbol{#1}}%
}

\begin{document}
\title{Probabilistic Integration of Object Level Annotations in \\Chest X-ray Classification}

\author[1]{Tom van Sonsbeek}
\author[1,2]{Xiantong Zhen}
\author[2]{Dwarikanath Mahapatra}
\author[1]{Marcel Worring}
\affil[1]{University of Amsterdam, Amsterdam, the Netherlands}
\affil[2]{Inception Institute of Artificial Intelligence, Abu Dhabi, UAE}

\newcommand{\x}{\mathbf{x}}
\newcommand{\y}{\mathbf{y}}
\newcommand{\z}{\mathbf{z}}
\newcommand{\cc}{\mathbf{c}}
\maketitle

\begin{abstract}
Medical image datasets and their annotations are not growing as fast as their equivalents in the general domain. This makes translation from the newest, more data-intensive methods that have made a large impact on the vision field increasingly more difficult and less efficient. In this paper, we propose a new probabilistic latent variable model for disease classification in chest X-ray images. Specifically we consider chest X-ray datasets that contain global disease labels, and for a smaller subset contain object level expert annotations in the form of eye gaze patterns and disease bounding boxes. We propose a two-stage optimization algorithm which is able to handle these different label granularities through a single training pipeline in a two-stage manner. In our pipeline global dataset features are learned in the lower level layers of the model. The specific details and nuances in the fine-grained expert object-level annotations are learned in the final layers of the model using a knowledge distillation method inspired by conditional variational inference. Subsequently, model weights are frozen to guide this learning process and prevent overfitting on the smaller richly annotated data subsets. The proposed method yields consistent classification improvement across different backbones on the common benchmark datasets Chest X-ray14 and MIMIC-CXR. This shows how two-stage learning of labels from coarse to fine-grained, in particular with object level annotations, is an effective method for more optimal annotation usage.
\end{abstract}

\section{Introduction}

The recent big advances in vision can be attributed to two main factors: algorithmic innovation and large amounts of data. Especially the availability of well-annotated datasets is showing to be a decisive factor \cite{tolstikhin2021mlp,dosovitskiy2020image,zhai2022scaling}. This is a noteworthy development when looking at the role the general vision domain has had on the medical imaging domain in recent years. 


Applying the early deep learning based generic computer vision solutions in the medical domain showed to work well, concretely: 1) The introduction of CNNs led to the first deep learning methods which outperformed radiologists\cite{rajpurkar2017chexnet}. 2) Although there exists an obvious gap between generic and medical data  \cite{raghu2019transfusion,alzubaidi2021novel} fine-tuning pre-trained models from the general vision domain as a basis of new models on medical data gives good results \cite{morid2021scoping,cheplygina2019cats}. 
\begin{figure}[!t]
    \centering
    \includegraphics[width=\linewidth]{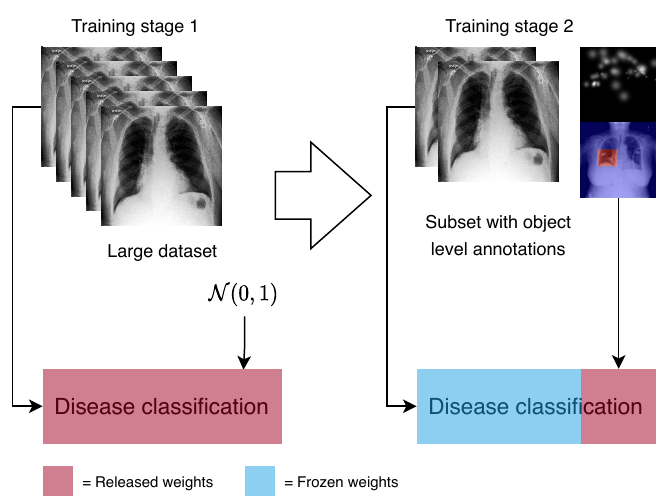}
    \caption{Model architecture overview for probabilistic integration of object level annotations.}
    \label{fig:my_label}
\end{figure}

A remaining challenge is whether the data-driven breakthroughs in the general vision domain can be translated to the medical domain in a similar way. In the general domain it is possible to increase annotated dataset sizes through crowdsourcing, web scraping, and label prediction through sophisticated vision-language methods. This allows for powerful models like the Vision Transformer~\cite{dosovitskiy2020image} and many methods building and innovating based on that \cite{liu2022swin,chen2021crossvit,Yuan_2021_ICCV}. Adaptations of these methods are leading to competitive results in the medical domain \cite{taslimi2022swinchex, he2022transformers}. However compared to previous methods their effect is not as large $(\sim1\%)$ as seen in the general domain $(\sim3\%)$. This can possibly be attributed to domain shift, scarcity of (annotated) data, and annotation quality.

Getting large scale annotations is more complicated in the medical domain, where reliable annotations originate from trained medical experts~\cite{zhou2021review}. In practise this means that current public medical datasets are small, and fully annotated, or large, and only partly annotated. Automatic sourcing of annotations through metadata or other information sources is an option when it is not possible to obtain expert level annotations on an entire dataset, even though this leads to a lower annotation quality \cite{zhou2019review,pooch2020can}. 

This shortage of expert level annotations is especially relevant for chest X-ray scans. The chest X-ray is one of the most common medical imaging modalities. The low cost, non-invasive nature and low patient impact explain their use as a primary diagnostic tool. The high volume of scans makes applying automated deep learning methods an interesting avenue that has been explored over the last years. As a consequence, there are multiple publicly available datasets exceeding 100k scans. Annotations on these datasets are often limited to global disease labels extracted from their corresponding radiology reports using Natural Language Processing (NLP)~\cite{wang2017chest8,johnson2019mimic}. Usage of radiology reports offers annotations for the entire dataset, but is also prone to unintended mislabeling and biases ~\cite{seyyed2021underdiagnosis}. Access to higher quality expert-level annotations can contribute to improved accuracies. 

Next to global disease labels, many x-ray datasets contain more precise object level annotations, made by clinicians for a subset of the data. At this stage these are: 1) bounding box annotations \cite{wang2017chest8,lanfredi2022reflacx}, describing the region of interest (ROI) within the X-ray in which signs of the disease are located. 2) eye gaze information \cite{karargyris2021creation,lanfredi2022reflacx}, the extracted gaze pattern of clinicians, tracked while they analyze and report on a chest X-ray scan using specialized software. These eye gaze maps contain valuable insights regarding the ROI and the analysis process of clinicians, since they show the exact locations that contribute to the decision making and reporting of the expert annotator. So far these object level annotations have not been widely used for improvement of classification performance. Bounding box information has been used to verify the location-awareness of classification methods. The recently introduced public eye gaze datasets have limited exploration so far. 

To improve disease classification, we see an opportunity for a scenario where we use a large dataset with global disease labels together with a smaller subset which contains more rich object level annotations. By doing so we encounter two challenges. First, the method should be able to learn from a dataset containing different granularities of labels, namely global disease labels and smaller subsets of eye gaze information maps from clinicians and disease bounding box annotations. Secondly, when training a deep neural network with small amounts of data, there is a risk of overfitting and loss of generalization to large datasets. We propose a two-stage optimization algorithm to incorporate object level annotations into representation learning of chest X-ray images. This unified training strategy can integrate different types of image labels. In this paper we make the following contributions: 

\begin{itemize}
    \item We propose a new probabilistic latent variable model for disease classification which is able to learn image representations by leveraging annotation of different granularities.
    \item We propose a two-stage optimization strategy enabling the model to learn low level features with large base datasets and more relevant features to diseases by integrating object level annotations.  
    \item We conduct extensive experiments which show that by combining the variational model and the two-stage training strategy it is possible to consistently improve disease classification performance, across two chest X-ray classification benchmark datasets, by over $3\%$. 
\end{itemize}

\begin{figure*}
    \centering
    \includegraphics[width=0.96\linewidth]{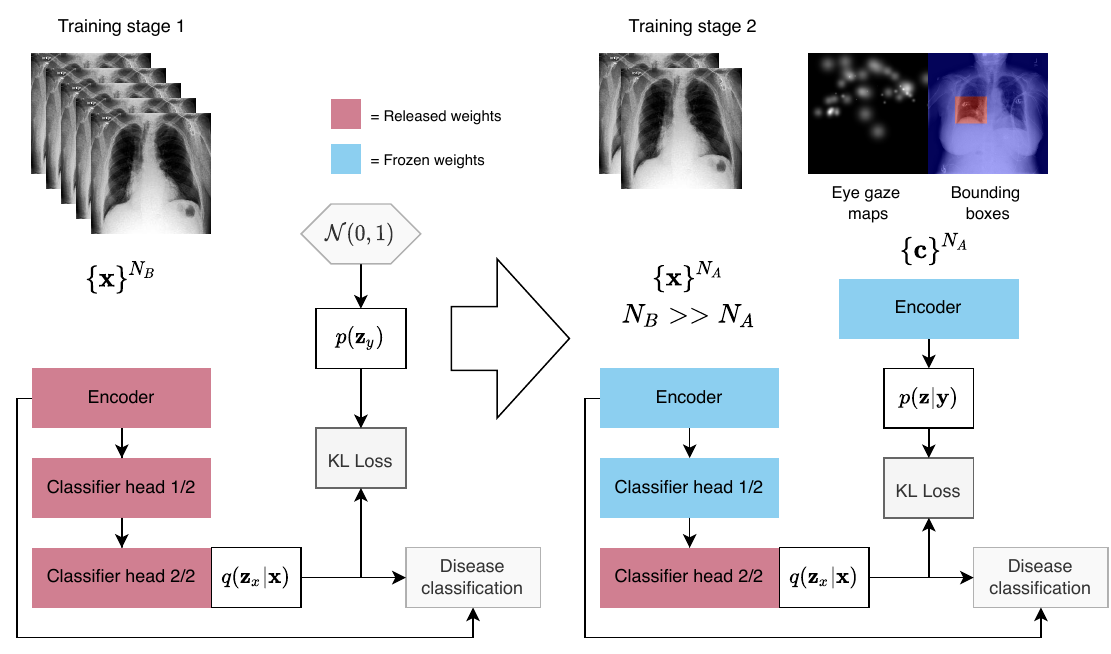}
    \caption{Architecture of probabilistic integration of object level annotations in chest X-ray classification. The architecture is composed of two learning stages: At first the entire model is trained on a large scale base dataset. Secondly fine-grained object level annotations on a data subset are infused through the variational prior, enabling better disease classification performance. }
    \label{fig:main}
\end{figure*}
\section{Related works} 

\paragraph{Chest X-ray classification}
Strides have been made on the problem of disease classification from chest X-rays over the last years. Since the introduction of ChexNet \cite{rajpurkar2017chexnet}, many new solutions for this classification problem have been provided, pushing the field forward. Methods range from supervised\cite{yao2018weakly,taslimi2022swinchex,pooch2020can,seyyed2020chexclusion,kim2021xprotonet}, to semi-supervised learning~\cite{liu2020semi,liu2021self}. A number of these methods use object level annotations to check the location-awareness of disease classification \cite{Liu_2019_ICCV,wu2020automatic, rajpurkar2020chexpedition,yan2018weakly,guendel2018learning}. Li \textit{et al.}~\cite{li2018thoracic} show that disease localization on large X-ray datasets can be aided by a subset of bounding box annotations. To the best of our knowledge, no existing methods are using object level annotations with the objective of enhancing classification performance on large scale X-ray datasets. 

\paragraph{Incorporation of eye gaze information}
This year, two datasets containing eye gaze information of clinicians analyzing chest X-ray scans were released. Huang \textit{et al.}~\cite{huang2021leveraging} showed that even a small scale eye gaze dataset has the ability to improve disease classification performance. Zhu \textit{et al.}~\cite{zhu2022multi} had similar findings and furthermore demonstrated that gaze information can also be used to generate more useful attention/saliency maps. These findings are promising and show that even with small amounts of data, eye gaze maps contain valuable information that improve disease classification results. It is still an open challenge for these methods to generalize to larger datasets, since these models are trained and tested on $<1\%$ of available chest X-ray images of the dataset they originate from. 

\paragraph{Probabilistic latent variable models}
A major successful application area of probabilistic latent variable models is in the multi-modal domain, due to the inherent versatility of using latent variables across domains. Examples of recent successful applications are in cross-modal retrieval~\cite{Chun_2021_CVPR} and multi-modal pose generation~\cite{Kundu_2020_WACV}. The use cases of probabilistic models also extend to the medical domain. For example in multi-modal segmentation of brain MRI scans \cite{hamghalam2021modality} and abdominal CT scans \cite{zhang2021modality}. Furthermore, multi-modal methods regarding chest X-rays, like report generation \cite{najdenkoska2021variational} and image-text disease classification~\cite{van2021variational} are based on probabilistic modelling methods. When we consider object level annotations as an additional data modality, the use of probabilistic latent variable models could be suitable.  

\section{Methodology}
Given a chest X-ray image, we would like to classify it into different categories of diseases. We frame disease classification based on chest X-ray images as a conditional variational inference problem by defining a probabilistic latent variable model. The latent variable is defined as the feature representation of the chest X-ray image. We will first give the preliminaries on variational auto-encoders, based on which we introduce the conditional inference model by designing a conditional prior. After that we will describe our two stage optimization method as shown in Fig.~(\ref{fig:main}).

\subsection{Preliminaries}
The variational auto-Encoder (VAE) \cite{kingma2013auto,rezende2014stochastic} is a probabilistic generative model which has been successful in many applications. The VAE has shown to be effective in learning low-dimensional representations of images. 

Specifically, given an input image $\x$ from a data distribution $p(\x)$, we would like to learn its n-dimensional vector representation $\z$ in the latent space to infer the posterior $p(\z|\x)$ over the latent variable $\z$. However, it is intractable to directly infer the posterior. Instead, we introduce a variational posterior $q(\z|\x)$ to approximate $p(\z|\x)$ by minimizing the KL divergence between them:
\begin{equation}
   D_{\rm{KL}}[q(\z|\x)||p(\z|\x)]
   \label{KL}
\end{equation}

By applying the Bayes' rule to Eq.~(\ref{KL}), we obtain the well-known evidence lower bound (ELBO):
\begin{equation}
    \mathcal{L}_{\rm{VAE}} = \mathbb{E}[\log p(\mathbf{x}|\mathbf{z})] - D_{\rm{KL}}[q(\mathbf{z}|\mathbf{x})||p(\mathbf{z})],
    \label{elbo}
\end{equation}
in which the posterior $q(\mathbf{z}|\mathbf{x})$ is dependent on $\x$ and the prior $p(\z)$ generally is assumed to be an isotropic Gaussian distribution $\mathcal{N}(0,I)$ over $\z$. 

The VAE is primarily used for generative modeling, while in this work, we would like to conduct supervised learning for disease classification. To do so, we design a new objective based on Eq.~(\ref{elbo}) by replacing the data likelihood with a conditional likelihood $p(\y|\z,\x)$. This gives rise to the objective function for supervised learning as follows:
\begin{equation}
    \mathcal{L} = \mathbb{E}[\log p(\y|\z,\x)] - \beta D_{\rm{KL}}[q(\z|\x)||p(\z)].
    \label{eq:base}
\end{equation}
where $\y$ is the disease label corresponding to the input image $\x$ and $\beta$ is the hyperparameter to control the weight of the KL term which is regarded as a regularizer. Intuitively, the objective in Eq.~(\ref{eq:base}) encourages the model to learn a compact latent representation of the input image $\x$ to maximally predict its disease label.


\subsection{Conditioning on Object Level Annotations}
The prior in Eq.~(\ref{eq:base}) is non-informative which serves to remove the redundancy in the latent representation. However, we would like the prior to be able to incorporate prior knowledge in order to achieve more informative representations. To this end, we propose to design a conditional distribution for the prior by leveraging extra label information provided by object level annotations, in particular bounding boxes and eye gaze maps. 

To be more specific, for a given image $\x$ in a data subset, we also have an object level annotation $\cc$ associated with it, which has identical dimensions as $\x$ . By specifying the prior as $p(\z|\cc)$ that is conditioned on the gaze map, we obtain a new objective function as follows: 
\begin{equation}
    \mathcal{L}_{\rm{SUB}} = \mathbb{E}[\log p(\y|\z,\x)] -\beta D_{\rm{KL}}[q(\z|\x)||p(\z|\cc)],
    \label{eq:sub}
\end{equation}
 The knowledge contained in the gaze map is used through the data-dependent prior by minimizing the KL term, which guides the model to extract the most relevant features from the input image. This objective provides a principled formalism to incorporate prior knowledge into representation learning.

\subsection{Two-stage optimization}
In general, we could directly use the objective in Eq.~(\ref{eq:sub}) to learn feature representations. In practice the extra annotations are not available for large scale datasets that usually only contain the global label annotations. The object level annotation, e.g., the gaze maps, are only provided for small scale datasets. In this case, we would not be able train a deep model from scratch on those small datasets.

In order to leverage both large scale datasets and also the annotations in small subsets of them, we propose a two-stage optimization. Our objective functions of variational inference provides the flexibility to do so. We consider a set of images $\{\x\}_{B}^{N_B}$, with global labels $\cc_B$ as a large base dataset which  contains only the disease labels. This dataset will be used for pre-training in stage 1. Furthermore we define a subset $\{\x\}_{S}^{N_S}$ $\in$ $\{\x\}_{B}^{N_B}$ with $N_B >> N_S$. It contains both additional object level annotations $\y_{S}$ and also global labels $\cc_{S}$. This dataset will be used for fine-tuning in Stage 2. The learning stages are designed to optimally utilize the existing label space.

\paragraph{Stage 1: Pre-training} In this stage, we train the model on the large scale base dataset by using the optimization objective in Eq.~(\ref{eq:base}).  We define the posterior as $q(\z_B|\x_B)$ conditioned on the input image $\x_B$ from the base datasets and the prior $p(\z)$ as a normal Gaussian distribution. Leveraging a large dataset during this stage leads to a comprehensive extraction of relevant image features. 

\paragraph{Stage 2: Fine-tuning} In this stage, we further fine-tune the model on the small dataset with object level annotations. We train the model with the optimization objective in Eq.~(\ref{eq:sub}). In this case, the prior is defined as $p(\z_S|\cc)$. This final stage allows for an optimal contextualization of image features already learnt in the earlier layers of the model. To combat overfitting and preserve generalization to the base dataset, model weights are frozen in earlier model layers.

\subsection{Qualitative evaluation}
The reasoning behind this method is that the inclusion of object-level annotations from clinicians can contribute to the model processing the image spatially in a similar way as clinicians do. The last step in which pixel level clinician annotations are infused is meant to accentuate certain features and locations in the image that are important according to the eye gaze and bounding box maps. This should be visible by comparing class activation maps (CAMs) of images from the base dataset before and after the last training step is applied. We define the following metric measuring this, with $S(\cdot)$ being a similarity metric like Mean Squared Error (MSE) or Dice: 
\begin{equation}
    {\Delta}S = \frac{S(\cc,CAM(\x)_{BASE})}{S(\cc,CAM(\x)_{SUB})}*100
\end{equation}
\subsection{Implementation}

The implementation of this method (Fig.~(\ref{fig:main})) is done with deep neural networks by adopting the amortization technique \cite{kingma2013auto}. Both the variational posterior $q(\z|\x)$ and the priors $p(\z|\cc),p(\z)$ are parameterized as fully factorized Gaussian distributions. 
The reparameterization trick~\cite{kingma2013auto} enables sampling from these distributions: $\mathbf{z}$: $\z^{(\ell)} = f(\x,\epsilon^{(\ell)})$ with $\epsilon^{(\ell)} \sim \mathcal{N}(0,I)$, and $f(\cdot)$ as a deterministic differentiable function.

In the posterior $q(\z|\x)$, $\x$ is considered to be the CNN representation of an X-ray image. In a similar fashion, $\cc$ in the prior $p(\z|\cc)$ is a CNN representation of an object level annotation, which could either be a gaze map or bounding boxes. Both prior and posterior are inferred by a multi-layer perceptron (MLP).

\section{Data and experimental settings}
\subsection{Datasets}
The proposed method is evaluated on two large public chest X-ray datasets which also contain smaller subsets of object level annotations by clinicians.  
\subsubsection{Chest X-ray14}
This dataset consists of $113,120$ frontal chest X-ray scans~\cite{wang2017chest8} of size $1024\times1024$ with 14 disease labels sourced from the accompanying radiology reports (which are not public) through a rule-based extraction method. 
A subset of bounding box annotations is available for this dataset. These are $1600$ disease bounding boxes distributed over $983$ X-ray scans. However, all of these annotations are contained in the test set. Using these annotations means that results can not be reported on the official test set. Instead, five-fold cross validation is applied, similarly adopted by Li \textit{et al.}~\cite{li2018thoracic} to solve this issue. We acknowledge that this approach makes comparability with previous methods that evaluate on the entire test set less reliable, since there is a mismatch in the assignment of train and test datasets. However, since this dataset serves as the major benchmark for disease classification of chest X-rays in recent years it is still included. 
\subsubsection{MIMIC-CXR}
This currently largest chest X-ray dataset contains $377,110$ images (sized $~2500\times3000$) which are a combination of frontal and sagittal views for $227,827$ studies in total. Labels are extracted through a process similar to Chest X-ray14, albeit with a slightly different label space~\cite{irvin2019chexpert}.

MIMIC-CXR has three different subsets that can produce fine-grained pixel level expert annotations. They originate from REFLACX ~\cite{lanfredi2022reflacx} and EGD-CXR ~\cite{karargyris2021creation}. REFLACX contains eye gaze information for $2616$ X-ray scans. Additionally, for the same subset, disease bounding boxes (BB) are also provided. 
EGD-CXR contains eye gaze maps for $1083$ X-ray scans. Since these annotations are spread across train and test sets results can be reported on the official test set.  

\subsection{Experimental settings}
X-ray images are standardized by normalization and rescaling to size $224\times224$ with center-cropping. This is in adherence with standards in this field, despite causing a slight performance drop~\cite{haque2021effect}. To measure consistency of our method we evaluate our experiments on commonly used CNN backbones, which are pre-trained on ImageNet~\cite{He_2016_CVPR}. The CNN backbones are: VGG16~\cite{vggsimonyan2014very}, ResNet50~\cite{reshe2016deep} and DenseNet121~\cite{huang2017densely}. Recent works showed the latter works best on Chest X-ray images \cite{rajpurkar2017chexnet, xue2019improved}. Finetuning is applied on the CNN backbone \cite{ke2021chextransfer,raghu2019transfusion}. 

Posterior $q(\z|\x)$ is inferred by two sequential two-layer MLPs with hidden dimension $512$. Note that the weights of the first MLP will be kept frozen during the conditioning on object level annotations in training stage 2. Prior $p(\z|\cc)$  is similarly generated through a two-layer MLP with $512$ hidden dimensions. 

The object level annotations (bounding boxes, eye gaze maps) are incorporated as pixel level annotations. They are represented as images with values ranging between 0 and 1. These should represent a pseudo-segmentation map of the crucial regions of the image. These maps will be passed through an ImageNet pre-trained CNN encoder of the same type as the X-ray image encoder. 

Eye gaze datasets contain fixations points of the radiologist on specific coordinates within the image, combined with how long these fixations lasted. Each fixation is characterized by a Gaussian with radius depending on the fixation's length in seconds (with empirical multiplier $a_{\sigma GAZE} = 10$). Bounding boxes are similarly mapped to the shape of the original X-ray images. To prevent edge issues with the CNN encoder, a Gaussian smoothing of the edges is applied with $\sigma_{BB}$ = 5. 

Grad-CAM~\cite{selvaraju2017grad} is used to compute CAMs. Training was done with one Ryzen 2990WX CPU and one NVIDIA RTX 2080ti GPU, using Adam \cite{kingma2014adam} optimization and early stopping with a tolerance of 1\%.

\section{Results and discussion}

\subsection{Improving disease classification}

\renewcommand\arraystretch{0.95}
\begin{table}[b]
\centering
\resizebox{\linewidth}{!}{%
\begin{tabular}{lllcc}\toprule
                   & Backbone    & Setting                              & AUC   & F1   \\ \midrule
Chest X-ray14 &             &                                      &       &      \\ \cmidrule(r){1-1}\cmidrule(rl){2-2}\cmidrule(rl){3-3}\cmidrule(l){4-5} 
\multicolumn{2}{l}{Semi-supervised}&\multicolumn{3}{l}{}\\ \cmidrule(r){1-1}
Aviles \textit{et al.}~\cite{aviles2019graphx}        & Graph & -                                    & 0.789 & -     \\
Liu \textit{et al.}~\cite{liu2020semi}        & DenseNet169 & -                                    & 0.792 & -     \\
Liu \textit{et al.}~\cite{liu2021self}        & DenseNet169 & -                                    & 0.811 & -     \\
\cmidrule(r){1-1}\cmidrule(rl){2-2}\cmidrule(rl){3-3}\cmidrule(l){4-5} 
\multicolumn{2}{l}{Supervised}&\multicolumn{3}{l}{}\\ \cmidrule(r){1-1}

Wang \textit{et al.}~\cite{wang2017chest8}        & ResNet50 & -                                    & 0.745 & -     \\
Yao \textit{et al.}~\cite{yao2018weakly}         & DenseNet121 & -                                    & 0.761 & -     \\
Guendel \textit{et al.}~\cite{guendel2018learning}     & DenseNet121 & -                                    & 0.807 &-      \\
Kim \textit{et al.}~\cite{kim2021xprotonet}& DenseNet121 & -                                    & \textBF{0.820} &-      \\
ViT~\cite{taslimi2022swinchex}              & Transformer & -                                    & 0.779 & -     \\
Taslimi \textit{et al.}~\cite{taslimi2022swinchex}           & Transformer & -                                    & 0.810  &-      \\ 
Li \textit{et al.}~\cite{li2018thoracic}$\footnotemark$     & ResNet50 & Base model                                    & 0.746 & -     \\
& & $+$Bounding boxes                                    & 0.797 & -     \\\cmidrule(r){1-1}\cmidrule(rl){2-2}\cmidrule(rl){3-3}\cmidrule(l){4-5}
Ours$^1$            & VGG16       & Base model                           & 0.754  & 0.24 \\
                   &             & $$+$$Bounding boxes                      & 0.786  & 0.25     \\ \cmidrule(r){2-2}\cmidrule(rl){3-3}\cmidrule(l){4-5} 
                   & ResNet50    & Base model                           & 0.763  & 0.24 \\
                   &             & $$+$$Bounding boxes                      & 0.793  & \textBF{0.26}     \\ \cmidrule(r){2-2}\cmidrule(rl){3-3}\cmidrule(l){4-5} 
                   & DenseNet121 & Base model                           & 0.772  & 0.24 \\
                   &             & $$+$$Bounding boxes                      & 0.809  & 0.25     \\ \midrule
MIMIC-CXR          &             &                                      &       &      \\ \cmidrule(r){1-1}\cmidrule(rl){2-2}\cmidrule(rl){3-3}\cmidrule(l){4-5} 
Pooch \textit{et al.}~\cite{pooch2020can}& DenseNet121 & -                                    & 0.828 &-      \\
Seyyed \textit{et al.}~\cite{seyyed2020chexclusion}& DenseNet121 & -                                    & 0.834 &-      \\
\cmidrule(r){1-1}\cmidrule(rl){2-2}\cmidrule(rl){3-3}\cmidrule(l){4-5} 
Ours               & VGG16       & Base model                           & 0.806 & 0.24 \\
                   &             & $+$BB - REFLACX                       & 0.814 & 0.25 \\
                   &             & $+$Gaze - REFLACX                      & 0.831 & 0.26 \\
                   &             & $+$Gaze - EGD-CXR                      & 0.827 & 0.26 \\
                   &             & \begin{tabular}[c]{@{}l@{}}$+$Gaze - EGD-CXR \&\\ BB - REFLACX\end{tabular}                      & 0.829 & 0.26 \\
                    &             & \begin{tabular}[c]{@{}l@{}}$+$Gaze - EGD-CXR \&\\ Gaze - REFLACX\end{tabular}                     & 0.811 & 0.24 \\\cmidrule(r){2-2}\cmidrule(rl){3-3}\cmidrule(l){4-5} 
                   & ResNet50    & Base model                           & 0.804 & 0.24 \\
                   &             & $+$BB - REFLACX                     & 0.813 & 0.25 \\
                   &             & $+$Gaze - REFLACX                      & 0.831 & 0.25 \\
                   &             & $+$Gaze - EGD-CXR                      & 0.834 & 0.26 \\
                   &             & \begin{tabular}[c]{@{}l@{}}$+$Gaze - EGD-CXR \&\\ BB - REFLACX\end{tabular}                   & 0.809 & 0.25 \\
                   &             & \begin{tabular}[c]{@{}l@{}}$+$Gaze - EGD-CXR \&\\ Gaze - REFLACX\end{tabular}                    & 0.832 & 0.26 \\                   
                   \cmidrule(r){2-2}\cmidrule(rl){3-3}\cmidrule(l){4-5}
                   & DenseNet121 & Base model                           & 0.807 & 0.25 \\
                   &             & $+$BB - REFLACX                      & 0.821 & 0.26 \\
                   &             & $+$Gaze - REFLACX                      & 0.827 & \textBF{0.27} \\
                   &             & $+$Gaze - EGD-CXR                      & \textBF{0.836} & \textBF{0.27} \\
                   &             & \begin{tabular}[c]{@{}l@{}}$+$Gaze - EGD-CXR \&\\ BB - REFLACX\end{tabular}                     & 0.815 & 0.25 \\
                   &             & \begin{tabular}[c]{@{}l@{}}$+$Gaze - EGD-CXR \&\\ Gaze - REFLACX\end{tabular}                     & 0.835 & \textBF{0.27} \\\bottomrule                   
\end{tabular}}
\caption{Disease classification performance of proposed method in AUC and F1 scores. Base model scores indicate the performance after the training stage I with a large base dataset. Other scores indicate the performance on the base dataset test set after the integration of object level annotations subsets in training stage 2.}

\label{tab:main}
\end{table}

The performance of our method across different CNN backbones is shown in Table~\ref{tab:main}.  We see that the addition of object level annotations improves classification results compared to base model results. This consistent improvement compared to the baseline model is the main feature of our method. Additionally, it shows to be competitive with prior works, performing better or within $\sim1\%$ in AUC score for Chest X-ray14, and reaching best scores on MIMIC-CXR. In this comparison, however, the difference in train/test split on Chest X-ray14 should be taken in consideration.  

\begin{table}[!t]
\centering
\resizebox{\linewidth}{!}{%
\begin{tabular}{llll}\toprule
           & Setting                              & $\Delta$MSE $(\%)$ & $\Delta$Dice $(\%)$ \\ \midrule
Chest X-ray14 &                                      &           &            \\ 
\cmidrule(r){1-1}\cmidrule(rl){2-2} \cmidrule(l){3-4}
VGG16              & Base model                           & -         & -          \\
                   & +Bounding boxes                      & +14         & +5          \\  \cmidrule(r){1-1}\cmidrule(rl){2-2} \cmidrule(l){3-4}
ResNet50           & Base model                           & -         & -          \\
                   & +Bounding boxes                      & +15        & +8          \\  \cmidrule(r){1-1}\cmidrule(rl){2-2} \cmidrule(l){3-4}
DenseNet121        & Base model                           & -         & -          \\
                   & +Bounding boxes                      & +6         & +7          \\  \midrule
MIMIC-CXR          &                                      &           &            \\ \cmidrule(r){1-1}\cmidrule(rl){2-2} \cmidrule(l){3-4}
VGG16              & Base model                           & -         & -          \\
                   & +Bounding boxes                      & +7         & +4          \\
                   & +Gaze - REFLACX                      & +12        & +10          \\
                   & +Gaze - EGD-CXR                      & +13        & +9        \\ 
                   & \begin{tabular}[c]{@{}l@{}}+Gaze - EGD-CXR \&\\ BB - REFLACX\end{tabular}                      & +3        & +4         \\
                    & \begin{tabular}[c]{@{}l@{}}+Gaze - EGD-CXR \&\\ Gaze - REFLACX\end{tabular}                      & +13        & +9         \\  \cmidrule(r){1-1}\cmidrule(rl){2-2} \cmidrule(l){3-4}
ResNet50           & Base model                           & -         & -          \\
                   & +Bounding boxes                      & +8         & +6          \\
                   & +Gaze - REFLACX                      & +15        & +9          \\
                   & +Gaze - EGD-CXR                      & +16        & +8          \\ 
                   & \begin{tabular}[c]{@{}l@{}}+Gaze - EGD-CXR \&\\ BB - REFLACX\end{tabular}                      & +7        & +6         \\
                    & \begin{tabular}[c]{@{}l@{}}+Gaze - EGD-CXR \&\\ Gaze - REFLACX\end{tabular}                      & +16        & +8         \\  \cmidrule(r){1-1}\cmidrule(rl){2-2} \cmidrule(l){3-4}
DenseNet121        & Base model                           & -         & -          \\
                   & +Bounding boxes                      & +8         & +5          \\
                   & +Gaze - REFLACX                      & +17        & +8          \\
                   & +Gaze - EGD-CXR                      & +15        & \textBF{+11}         \\
                   & \begin{tabular}[c]{@{}l@{}}+Gaze - EGD-CXR \&\\ BB - REFLACX\end{tabular}                      & \textBF{+19}       & +5         \\
                    & \begin{tabular}[c]{@{}l@{}}+Gaze - EGD-CXR \&\\ Gaze - REFLACX\end{tabular}                      & +14        & +7         \\\bottomrule      
\end{tabular}%
}
\caption{Difference in MSE and Dice similarity score between GRADCAM activation map and object level clinician annotation before and after fine-tuning with this subset of object level annotations.}
\label{tab:dcms}
\end{table}
\begin{table*}[!t]
\centering
\resizebox{0.90\textwidth}{!}{%
\begin{tabular}{ccccccccccc}\toprule
                   & \multicolumn{3}{c}{MIMIC-CXR subset}                                              & Chest X-ray14 subset  & \multicolumn{2}{c}{VGG16} & \multicolumn{2}{c}{ResNet50} & \multicolumn{2}{c}{Densenet121} \\ \cmidrule(rl){2-4} \cmidrule(rl){5-5}\cmidrule(rl){6-7}\cmidrule(rl){8-9}\cmidrule(l){10-11}
                   & REFLACX gaze              & REFLACX BB                & EGD Gaze                  & BB                        &  AUC          & F1         & AUC           & F1           & AUC             & F1            \\\cmidrule(r){1-1} \cmidrule(rl){2-4} \cmidrule(rl){5-5}\cmidrule(rl){6-7}\cmidrule(rl){8-9}\cmidrule(l){10-11}
Chest X-ray 14 & \xmark     & \xmark     & \xmark     & \xmark     & 0.754         & 0.24       & 0.763          & 0.25         & 0.772            & 0.24          \\
                   & \checkmark & \checkmark & \checkmark & \checkmark & 0.769         & 0.24       & 0.757          & 0.25         & 0.789            & 0.25          \\
                   & \checkmark & \checkmark & \checkmark & \xmark     & 0.737         & 0.24       & 0.731          & 0.22         & 0.766            & 0.24          \\
                   & \xmark     & \xmark     & \xmark     & \checkmark & \textBF{0.786}         & \textBF{0.25}       & \textBF{0.793}          & \textBF{0.26}         & \textBF{0.809}            & \textBF{0.26}          \\\cmidrule(r){1-1} \cmidrule(rl){2-4} \cmidrule(rl){5-5}\cmidrule(rl){6-7}\cmidrule(rl){8-9}\cmidrule(l){10-11}
MIMIC-CXR          & \xmark     & \xmark     & \xmark     & \xmark     & 0.806        & 0.24       & 0.804         & 0.24         & 0.807           & 0.25          \\
                   & \checkmark & \checkmark & \checkmark & \checkmark & 0.813        & 0.24       & 0.818         & \textBF{0.26}         & 0.826           & 0.25          \\
                   & \xmark     & \xmark     & \xmark     & \checkmark & 0.794        & 0.23       & 0.788         & 0.24         & 0.798           & 0.24          \\
                   & \checkmark & \checkmark & \checkmark & \xmark     & \textBF{0.827}        & \textBF{0.26}       & \textBF{0.834}         & \textBF{0.26}         & \textBF{0.836}           & \textBF{0.27}\\\bottomrule         
\end{tabular}%
}
\captionsetup{width=0.96\linewidth}
\caption{Cross-domain performance, for which annotation subsets (eye gaze maps and bounding boxes (BB)) are exchanged between datasets during the second training step. In this setting the base model is trained only on the source dataset (left column).}
\label{tab:cross}
\end{table*}
\begin{table*}[!t]
\centering
\resizebox{0.60\linewidth}{!}{%
\begin{tabular}{cccccccccc}\toprule&&&&\multicolumn{2}{c}{VGG16}&\multicolumn{2}{c}{ResNet50}&\multicolumn{2}{c}{Densenet121}\\
              & \multicolumn{1}{l}{Encoder} & CH1 & CH2 & AUC   & F1   & AUC   & F1   & AUC   & F1   \\\cmidrule(r){1-1} \cmidrule(rl){2-4} \cmidrule(rl){5-6}\cmidrule(rl){7-8}\cmidrule(l){9-10}
Chest X-ray14 & \xmark       & \checkmark   & \xmark   & 0.751  & 0.25 & 0.771  & 0.25 & 0.759  & 0.24 \\
              & \xmark       & \checkmark   & \checkmark   & 0.767  & 0.25 & 0.764  & \textBF{0.26} & 0.768  & \textBF{0.26}\\
              & \checkmark       & \checkmark   & \checkmark   & 0.723  & 0.25 & 0.722  & 0.24 & 0.723  & 0.23 \\
              & \xmark       & \xmark   & \checkmark   & \textBF{0.786}  & \textBF{0.26} & \textBF{0.793}  & \textBF{0.26} & \textBF{0.809}  & \textBF{0.26} \\\cmidrule(r){1-1} \cmidrule(rl){2-4} \cmidrule(rl){5-6}\cmidrule(rl){7-8}\cmidrule(l){9-10}
MIMIC-CXR     & \xmark       & \checkmark   & \xmark   & 0.779 & 0.19 & 0.765 & 0.17 & 0.782 & 0.16 \\
              & \xmark       & \checkmark   & \checkmark   & 0.783 & 0.20 & 0.788 & 0.17 & 0.792 & 0.18 \\
              & \checkmark       & \checkmark   & \checkmark   & 0.690 & 0.16 & 0.703 & 0.16 & 0.710 & 0.17 \\
              & \xmark       & \xmark   & \checkmark   & \textBF{0.827} & \textBF{0.26} & \textBF{0.834} & \textBF{0.26} & \textBF{0.836} & \textBF{0.27}\\\bottomrule
\end{tabular}%
}
\captionsetup{width=0.71\linewidth}
\caption{Ablation study over second training step. The checkmark indicates whether the model components' weights are released during the second training step. The model components (Fig.~\ref{fig:main}) are: encoder, classifier head (CH) 1 and 2. }
\label{tab:abl}
\end{table*}

The robustness and consistency of \footnotetext{Five-fold cross validation}the method are reflected in the effectiveness of our method over multiple CNN backbones and two different datasets. Results on MIMIC-CXR indicate that eye-gaze information is a more valuable object level annotation for our method than bounding boxes. A peculiar finding in Table~\ref{tab:main} is the performance difference between integration of REFLACX and EGD-CXR eye gaze maps. The sizes of these datasets are $\sim$3k and $\sim$1k respectively. It would be expected that the larger dataset would be more effective in leveraging higher classification scores, while the contrary is shown in the results. It can not be said with certainty to which dataset property this can be attributed to, but we can conclude that data quality in these object level annotations is an important factor. Simultaneous infusion of object annotation of both sources (EGD-CXR and REFLACX) was tested on MIMIC-CXR. The results reveal that that simultaneous integration of two eye gaze subsets yields better results than integration of a bounding box and eye gaze subset. Therefore we conclude that a certain consistency within the object level annotations is beneficial to reach optimal performance. Results split by disease class are listed in supplementary material A.    
\subsection{Grad-CAM similarity to object level annotations}

By using our method on integration of expert object level annotations, we guide our classification method to consider an X-ray image in a spatially similar way as a clinician. Table~\ref{tab:dcms} shows the difference in similarity between the CAM of an X-ray and their object level annotation after (1) the base dataset training and (2) after the subset training with object level annotations. A positive value means that the similarity to the object level annotations increased after training with the subset containing object level annotations. As the results in Table~\ref{tab:dcms} show an increase in these similarity scores, we confirm that the reasoning behind our methodology is valid. The increase in similarity scores is higher for eye gaze maps than for bounding boxes. This can be an indication that eye gaze maps are a more informative object level annotation than bounding boxes.

\begin{figure*}[!t]
    \centering
    \includegraphics[width=0.74\linewidth]{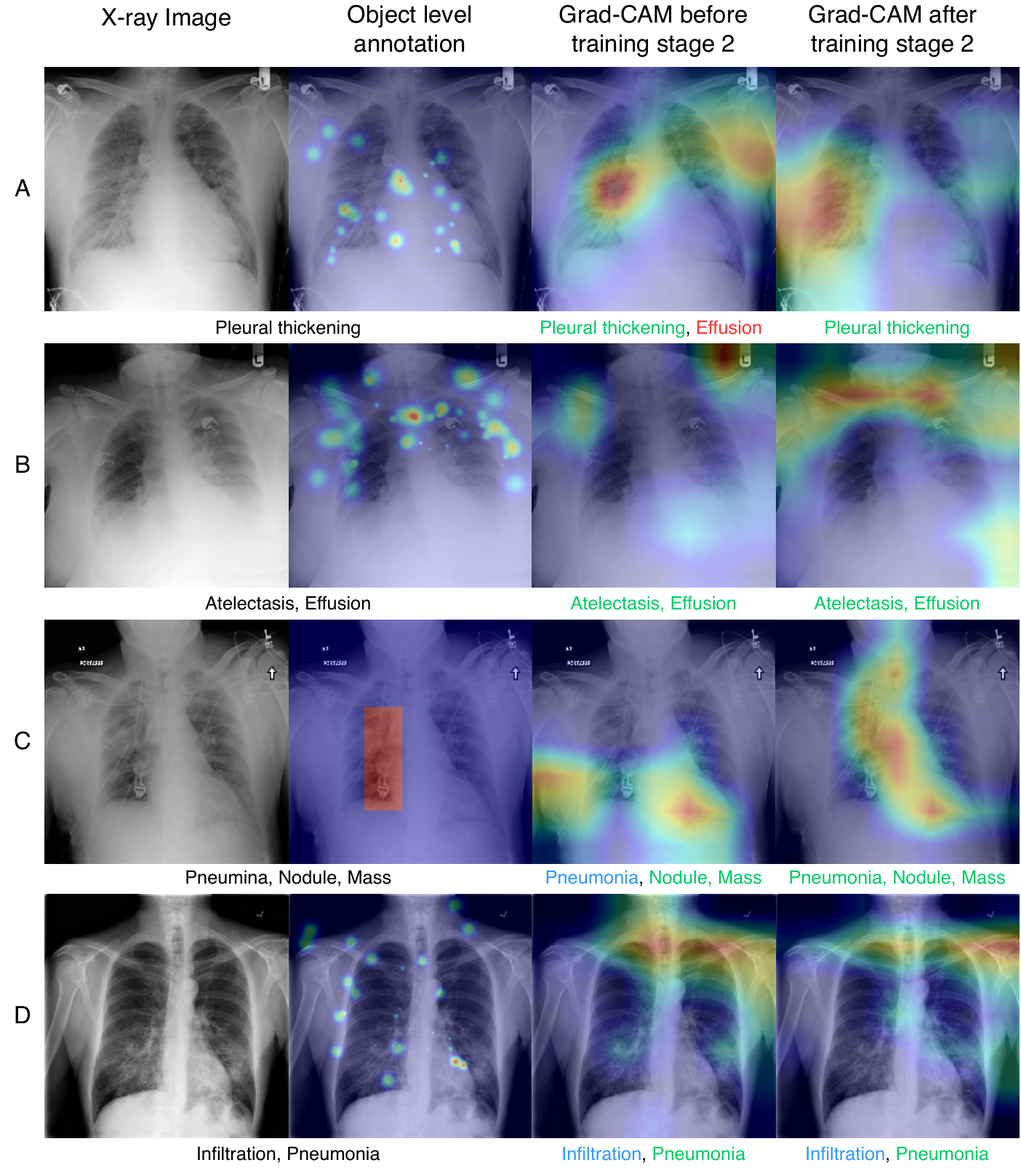}
    \captionsetup{width=0.90\linewidth}
    \caption{Visualisation of change in Grad-CAM before and after the second training stage with conditioning on object level annotations. Green, blue and red labels stand for correct, missed and wrong prediction respectively. (A) EGD-CXR eye gaze (B) REFLACX eye gaze (C) Chest X-ray14 bounding box (D) REFLACX eye gaze.}
    \label{fig:gcamview}
\end{figure*}

In Fig.~(\ref{fig:gcamview}) several examples are shown to illustrate how infusion of object level annotations can benefit classification capability. Fig.~(\ref{fig:gcamview}A) shows how the eye gaze pattern of the clinician has a bottom-left focus, which seems to be adopted through training stage 2. After this stage the model no longer incorrectly detects the 'Efusion' label.  In Fig.~(\ref{fig:gcamview}B-C) similar GradCAM shift pattern can be observed. With an unexpressive eye gaze map as in Fig.~(\ref{fig:gcamview}D) we see limited effects of conditioning on the gaze map. In this instance we can see that the missed 'Infiltration' label after stage 1 training is also not detected after stage 2 training.

\subsection{Cross-dataset learning}
Another interesting but challenging setting to evaluate our method on is when the object-level annotations used in the second training step originate from another similar dataset. The transferability and similarity within chest X-ray datasets has been studied in earlier works and indicated that  there is a non-negligible domain shift between current public chest X-ray datasets~\cite{raghu2019transfusion,ke2021chextransfer,pooch2020can,hosseinzadeh2021systematic}. Our results in Table~\ref{tab:cross} confirm this domain shift. Using the annotation subset of a different base dataset consistently does not lead to classification improvements.  Interestingly though, when including more out-of-dataset subsets, the performance compared to base dataset training can improve, while still being lower compared to in-dataset training. 

\subsection{Effect of layer freezing in two-stage training}

We show different settings with frozen or released weights over the final training step where we condition the model on object-level annotations in Table~\ref{tab:abl}. These experiments confirm the benefit of only updating the model weights in the last model layers when fine-tuning with a small data subset. Unfreezing the weights of the earlier layers leads to a steep decline in performance. Furthermore they show that fine-tuning of the last layers is needed to obtain the best classification results. 
An additional downside of unfreezing the earlier model layers is the faster occurrence of overfitting which is also detrimental for the generalization on the classification on the base dataset.

\section{Conclusion} 

In this paper we introduce a probabilistic latent variable model for classification of chest X-rays. It tackles the problem of label scarcity in the medical imaging domain. This model is able to process different types of label granularities, resulting in an efficient usage of all available labels. To achieve this, a two-stage method is introduced. In its first stage, disease classification on large base dataset is done to learn global image features. In the second stage, model weights in earlier layers are frozen. This enables conditioning on a small data subset of object level annotations in the form of eye gaze maps and bounding boxes for better contextualization of features learnt in the first training stage.  
This simple, yet effective, approach proves it effectiveness by consistently improving performance on multiple datasets. It provides an interesting outlook on how to manage data and how to utilise smaller subsets of rich annotated data in an effective manner. 

\section*{Acknowledgements}
This work is financially supported by the Inception Institute of Artificial Intelligence, the University of Amsterdam and the allowance Top consortia for Knowledge and Innovation (TKIs) from the Netherlands Ministry of Economic Affairs and Climate Policy.

{\small
\bibliographystyle{ieee_fullname}
\bibliography{egbib}
}

\end{document}